\documentclass{article}
\usepackage{spconf,amsmath,amssymb,graphicx}
\usepackage{times,booktabs,tabularx,array,multirow}
\usepackage{microtype,algorithm,algorithmic}
\usepackage{url,xurl,cite,placeins}
\usepackage[hidelinks,breaklinks=true,bookmarks=true]{hyperref}
\newcommand{\normF}[1]{\left\lVert #1\right\rVert_{\mathrm F}}
\newcommand{\normop}[1]{\left\lVert #1\right\rVert_{2}}
\newcommand{\sr}{\operatorname{sr}}
\newcommand{\Proj}{\mathcal P}
\newcommand{\NS}{\operatorname{NS}_{5}}
\newcommand{\clip}{\operatorname{clip}}
\newcommand{\method}{EnergyCal}
\newcommand{\leadpara}[1]{\par\smallskip\noindent\textbf{#1}\ }
\title{EnergyCal: Taming Loss Spikes in Muon-Family \\ Optimizers for Sparse MoE Training}
\name{Masato Fujitake}
\address{FA Research, Fast Accounting Co., Ltd.\\
{\tt\small fujitake@fastaccounting.co.jp}}
\hypersetup{pdftitle={EnergyCal: Taming Loss Spikes in Muon-Family Optimizers for Sparse MoE Training},pdfauthor={Masato Fujitake},pdfsubject={EnergyCal: Taming Loss Spikes in Muon-Family Optimizers for Sparse MoE Training}}
\begin{document}
\maketitle
\pagestyle{plain}
\begin{abstract}Sparse mixture-of-experts (MoE) language-model training can exhibit large loss spikes with Muon-family optimizers.
We introduce EnergyCal, a lightweight rule that attenuates routed-expert updates according to the normalized squared Frobenius norm of their finite-step Newton–Schulz directions.
EnergyCal leaves the spectral transform unchanged and adds no persistent optimizer state. We evaluate it on a 4B-parameter MoE model.
Against a development-selected, lower-learning-rate MuonH recipe, EnergyCal–MuonH reduces mean thresholded loss-spike burden by 99.51\% and final geometric-mean perplexity by 4.06\%, with a modest throughput cost.
With plain Muon at matched nominal learning rate and weight decay, the reduction in mean spike burden averages 95.98\% across the two tested exponents, accompanied by small endpoint improvements.
These findings support stabilization both with and without Hyperball in the evaluated setting. 
\end{abstract}
\begin{keywords}Muon, MuonH, mixture-of-experts, energy calibration, training stability\end{keywords}
\section{Introduction}\label{sec:intro}\begin{figure}[!t]
\centering\small
\includegraphics[width=\columnwidth]{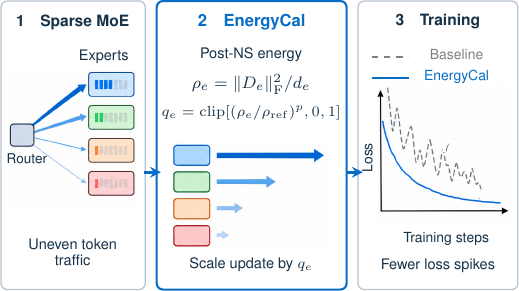}
\caption{\textbf{EnergyCal overview.} (1) Sparse routing creates uneven expert traffic. (2) Each routed matrix's post-NS energy determines $q_e$; when $q_e>0$, scaling preserves the individual NS direction, not necessarily the displacement after Hyperball projection or writeback. (3) Continuous calibration reduces observed thresholded loss-spike burden. 
Curves are schematic, not measurements; coefficients use direction energy, not token counts.}
\label{fig:energycal-overview}
\end{figure}

Sparse mixture-of-experts (MoE) language models route tokens unevenly across experts, creating time-varying optimization conditions~\cite{fujitake2026mesh}.
This setting can be challenging for Muon-family optimizers, which use Newton--Schulz (NS) iterations to transform momentum-based updates. 
Muon has shown promise for language-model pretraining~\cite{jordan2024muon,shah2025practical,liu2025scalable}, while Hyperball introduces explicit control over weight and update norms~\cite{wen2026hyperball}. 
Its Muon variant, MuonH, normalizes update directions and projects weights onto a fixed-radius sphere. In our custom 4B-parameter DeepSeekMoE-style model, both Muon and MuonH exhibit intermittent loss spikes, although all full-schedule runs complete. 
We investigate whether these loss excursions can be mitigated through a simple modification of routed-expert updates.

An expert can retain nonzero momentum even when no tokens are assigned to it in the current step.
This motivates examining expert-wise update magnitudes, particularly under MuonH's fixed-norm normalization. 
We propose \method{}, a lightweight rule that attenuates each routed-expert update according to the normalized squared Frobenius norm of its finite-step NS direction. 
Figure~\ref{fig:energycal-overview} illustrates the method. 
The coefficient is bounded between zero and one and is applied continuously during training.
\method{} leaves the NS transform, momentum, and router objective unchanged, requires no token-utilization statistics, and adds no persistent optimizer state. 
Its energy statistic is not scale-invariant stable rank: low energy can reflect direction concentration, amplitude, or both. 

We evaluate \method{} in two separate full-schedule studies, each using three initialization seeds. 
First, a frozen EnergyCal--MuonH recipe reduces mean thresholded loss-spike burden by 99.51\% and final geometric-mean perplexity by 4.06\% relative to a development-selected, lower-learning-rate MuonH recipe, with 1.62\% lower mean throughput. 
Second, applying \method{} to plain Muon at matched nominal learning rate and weight decay reduces mean spike burden by 95.98\% on average across the two tested exponents, with small endpoint improvements.
Together, these results show that stabilization is observed both with and without Hyperball in the evaluated MoE setting. However, plain Muon retains intermediate-quality and speed advantages in the first study, and different evaluation banks prevent a direct ranking of the calibrated optimizer families.

Our contribution is an empirical evaluation of one-sided, post-NS energy attenuation across MuonH and plain Muon. 
Unlike utilization- or alignment-guided expert scaling methods~\cite{shaier2026excitation,vaidya2026compass}, \method{} relies only on the energy of each logical matrix's NS direction. 
The results do not establish general optimizer superiority or identify the cause of individual spikes: single-update interventions fail to rescue two examined events, and the plain-Muon study lacks a routed-only uniform-scaling control to separate energy-dependent attenuation from a simpler reduction in update magnitude. 
We therefore focus on the measured stability benefits and their practical trade-offs within the tested setting; Section~\ref{sec:discussion} discusses these evidence boundaries.

\section{Related Work}\label{sec:related}\leadpara{Muon-family optimization and Hyperball.}
Muon applies a short Newton--Schulz (NS) iteration to momentum-based matrix updates~\cite{jordan2024muon}. Large-scale studies identify weight decay and update scaling as important for Muon pretraining~\cite{liu2025scalable}. Our plain-Muon baseline uses a shape factor of one rather than the full Moonlight recipe. Hyperball controls update and weight radii; its Muon variant, MuonH, normalizes trial directions and projects weights onto a fixed-radius sphere~\cite{wen2026hyperball}. EnergyCal changes routed-expert step magnitudes after NS without changing the spectral transform.

\leadpara{Expert-wise update scaling.}
Existing methods scale expert updates using other signals. Excitation uses expert utilization~\cite{shaier2026excitation}, while ExpertMuon-Compass uses post-orthogonalization gradient alignment across experts and rows~\cite{vaidya2026compass}. EnergyCal instead computes an independent coefficient $q\in[0,1]$ from each routed matrix's normalized squared post-NS Frobenius norm, without utilization inputs or cross-expert normalization. For $q>0$, scalar scaling preserves the NS direction; when $q=0$, the directional trial vanishes. Neither case guarantees the direction of the stored parameter displacement after Hyperball projection or low-precision writeback. Direction preservation itself is not unique to EnergyCal.

\leadpara{Magnitude and spectral corrections.}
NorMuon uses neuron-level normalization after orthogonalization~\cite{li2026normuon}, and TrasMuon uses RMS calibration and energy-based clipping~\cite{cheng2026trasmuon}. Thus, energy-aware scaling is not new by itself. Pion changes NS iterations~\cite{fan2026pion}, while SoftMuon uses smooth spectral relaxation~\cite{feoktistov2026softsign}. EnergyCal instead applies a matrix-level scalar after the unchanged NS transform. We do not report native-method head-to-head comparisons with these alternatives.

\leadpara{Training stability in sparse models.}
Training instability in sparse MoE models has motivated router-centered approaches, including ST-MoE's router z-loss and methods for reducing routing fluctuations~\cite{zoph2022stmoe, dai2022stablemoe}. 
Spike-aware optimization has also been studied in dense language-model training through momentum resets and gradient clipping~\cite{huang2025spam}.
Unlike these approaches, EnergyCal attenuates routed-expert updates using post-NS direction energy without modifying routing objectives or introducing spike-triggered interventions.

\leadpara{Finite-step orthogonalization.}
Recent analyses characterize the convergence and spectral behavior of Muon with finite-step Newton--Schulz iterations~\cite{kim2026convergence,magakyan2026spectral}.
MuonEq further investigates pre-NS equilibration to improve the geometry of orthogonalized updates~\cite{chang2026muoneq}.
EnergyCal instead leaves the NS transform unchanged and applies a matrix-level attenuation coefficient to its returned direction.

\leadpara{Evaluation and optimizer comparisons.}
Optimizer rankings depend on tuning and training budget~\cite{wen2026fantastic}; angular effective learning rates also complicate the interpretation of Hyperball's nominal learning rate~\cite{xiao2026free}. We therefore report final quality, intermediate quality, loss-spike burden, and throughput separately. Matched nominal learning rate does not imply matched parameter movement, and fixed-token quality is distinct from equal-wall-time quality.

\section{Energy-Calibrated Updates}\label{sec:method}
\subsection{Logical matrices and the base update}

Let $W\in\mathbb R^{m\times n}$ denote a logical parameter matrix, $G_t$ its gradient after accumulation, distributed reduction, and common clipping, and $B_t$ its momentum.
The evaluated implementation uses Nesterov momentum:
\begin{equation}
B_t=\beta B_{t-1}+G_t,\qquad
A_t=G_t+\beta B_t,
\end{equation}
with $\beta=0.95$.
A five-step quintic Newton--Schulz (NS) map produces the update direction $D_t=\NS(A_t)$.
NS arithmetic uses BF16, while momentum and norm calculations use FP32. 
Because the iteration is finite, $D_t$ is an approximate orthogonalized direction, not an exact polar factor. 
Implementation details and numerical guards are provided in Appendix~\ref{app:implementation}.

Gate and up projections share fused storage for efficient forward execution but are optimized as separate logical matrices by both MuonH and plain Muon. 
Each routed gate, up, and down matrix has $d=\min(m,n)=256$.
Attention and shared-expert matrices also receive matrix-optimizer updates, whereas embeddings/tied output weights, router weights, and vector parameters use the Adam fallback.

MuonH additionally constrains weight and update norms.
For the initial radius $R=\normF{W_0}$, define
$\Proj_R(X)=RX/\max(\normF{X},\epsilon_H)$.
The canonical MuonH update forms
\begin{align}
\Delta_t^{\rm trial}
&=-\eta_tR
\frac{D_t}{\max(\normF{D_t},\epsilon_H)},
\label{eq:trial}\\
\widetilde W_{t+1}
&=\Proj_R(W_t+\Delta_t^{\rm trial}),
\label{eq:project}
\end{align}
followed by BF16 parameter writeback using the common dithered precision policy. 
The trial update has a prescribed Frobenius length when its denominator guard is inactive, but the subsequent projection and writeback can change the resulting parameter displacement.
We therefore distinguish the trial displacement, FP32 projected chord, BF16 stored chord, and resulting function change. 
All MuonH-family comparisons use the same projection and writeback policy.

\subsection{Expert-wise energy calibration}

\method{} introduces a one-sided attenuation of routed-expert updates based on the energy of their
post-NS directions. 
For each logical matrix, define
\begin{align}
\rho_t
&=\frac{\normF{D_t}^{2}}{d},
\label{eq:energy}\\
q_t
&=\clip\!\left[
\left(\frac{\rho_t}{\rho_{\rm ref}}\right)^p,
q_{\min},1
\right].
\label{eq:coefficient}
\end{align}
Here, $\rho_t$ is the normalized squared Frobenius norm of the finite-step NS direction, $\rho_{\rm ref}$ is a reference energy, and $p$ controls the strength of attenuation. 
With $0\le q_{\min}\le1$ and $p>0$, the coefficient satisfies $q_t\in[0,1]$. 
For the evaluated choice $q_{\min}=0$, directions with $\rho_t<\rho_{\rm ref}$ receive $q_t<1$, while directions at or above the reference receive no additional attenuation from EnergyCal.

For MuonH, \method{} replaces $\eta_t$ in Eq.~\eqref{eq:trial} with $q_t\eta_t$ only for
routed-expert matrices. 
All other matrices retain $q_t=1$. As illustrated in Fig.~\ref{fig:energycal-overview}, positive scaling preserves the orientation of the individual NS direction, but not necessarily that of the projected or stored parameter displacement. 
Setting $q_t=0$ nulls the trial update, not necessarily the final stored parameter change. 
The figure uses expert indices $e$, whereas the equations use update indices $t$.

The coefficient is computed independently for each routed logical matrix at every update. 
It uses neither token-utilization statistics nor comparisons across experts. 
The NS transform, momentum, router objective, attention and shared-expert updates, and Adam fallback remain unchanged. 
No additional gradient history or persistent optimizer state is introduced. 
MuonH reuses its existing direction norm, whereas plain Muon requires an additional norm reduction.

The primary EnergyCal--MuonH recipe uses $p=1$, $\rho_{\rm ref}=0.8$, and $q_{\min}=0$, with calibration active from the first update and no sequence auxiliary loss. 
These settings were fixed during development and not reselected on the confirmation seeds. 
The energy statistic is used as an attenuation signal, not as a calibrated probability of update correctness or a validated signal-to-noise estimate.

\begin{algorithm}[t]
\caption{EnergyCal--MuonH on one logical matrix}
\label{alg:energy}
\begin{algorithmic}[1]
\REQUIRE $W,G,B,R,\eta,\beta,p,\rho_{\rm ref},q_{\min}$
\STATE $B\leftarrow\beta B+G$; $A\leftarrow G+\beta B$
\STATE $D\leftarrow\NS(A)$; $s\leftarrow\normF{D}$ in FP32
\STATE $q\leftarrow1$
\IF{$W$ is a routed-expert matrix}
\STATE $q\leftarrow\clip((s^2/(d\rho_{\rm ref}))^p,q_{\min},1)$
\ENDIF
\STATE $V\leftarrow W-\eta RqD/\max(s,\epsilon_H)$
\STATE $\widetilde W\leftarrow RV/\max(\normF{V},\epsilon_H)$
\STATE Write $\widetilde W$ using the shared precision policy
\end{algorithmic}
\end{algorithm}

\subsection{Transfer to plain Muon}

The same energy coefficient can be applied to plain Muon without Hyperball normalization or projection. 
The resulting update is
\begin{equation}
\widetilde W_{t+1}
=(1-\eta_t\lambda)W_t-\eta_t s_Wq_tD_t,
\label{eq:plain_energy}
\end{equation}
followed by the common writeback policy.
Here, $\lambda$ denotes decoupled weight decay and $s_W$ is the matrix-shape factor. 
All evaluated plain-Muon runs use $s_W=1$, unlike the scale-adjusted large-model recipe of Liu et al.~ \cite{liu2025scalable}.

As in MuonH, energy is measured independently for each routed gate, up, and down matrix before any
shape-factor scaling. 
Only routed matrices may have $q_t\ne1$; weight decay, momentum, non-routed updates, and the Adam fallback remain unchanged.
Unlike MuonH, plain Muon does not subsequently normalize the update to a prescribed Frobenius length or project the weights onto a sphere. 
Consequently, the same coefficient acts on a different base update and does not imply the same parameter displacement.
In particular, $q_t=0$ removes the NS update term but does not disable weight decay.

We evaluate $p\in\{0.5,1\}$ with $\rho_{\rm ref}=0.8$ and $q_{\min}=0$, without retuning the nominal learning rate. In this setting, \method{} adds energy-dependent attenuation to plain Muon's existing direction-magnitude dependence, rather than compensating for Hyperball normalization.
Whether this improves training stability is therefore evaluated empirically.

\subsection{Geometric motivation, not a causal theorem}

The motivation for post-NS energy calibration is partly geometric. For a nonzero direction $D$ and an inactive denominator guard, the MuonH trial update satisfies
\begin{equation}
\sr(D)=\frac{\normF{D}^2}{\normop{D}^2},
\qquad
\normop{\Delta^{\rm trial}}
=\frac{\eta Rq}{\sqrt{\sr(D)}}.
\label{eq:identity}
\end{equation}
Thus, at a fixed trial Frobenius length $\eta Rq$, a smaller stable rank corresponds to a larger
spectral concentration of the update. 
However, the operator norm remains bounded by
$\normop{\Delta^{\rm trial}}\le\eta Rq$.
This identity concerns the trial update, not the projected or BF16-stored displacement, and does not bound the nonlinear network's loss change or identify the cause of individual spikes.

Importantly, the energy used by \method{} is not stable rank. 
The two quantities satisfy
\begin{equation}
\rho(D)
=\frac{\sr(D)}{d}\normop{D}^{2}.
\label{eq:factorization}
\end{equation}
Unlike stable rank, $\rho(D)$ is not invariant under rescaling $D\mapsto aD$. 
Small energy can therefore arise from concentrated singular values, small direction amplitude, or both. Energy and normalized stable rank become approximately proportional only when the operator norm is independently known to remain approximately constant.

The exponent also changes how attenuation interacts with MuonH normalization.
In the unclipped, unguarded regime, $p=1/2$ gives
\begin{equation}
\frac{qD}{\normF{D}}
=\frac{D}{\sqrt{d\rho_{\rm ref}}},
\end{equation}
so the Frobenius normalization cancels algebraically. 
The trial operator norm still depends on $\normop{D}$, however, and this identity does not predict which exponent yields better language-model training outcomes.
The primary MuonH choice $p=1$ is empirical.

These relations motivate energy-dependent attenuation but do not establish a causal mechanism for loss spikes. 
In particular, Eq.~\eqref{eq:identity} does not describe the unnormalized plain-Muon update, and a zero MuonH trial is not equivalent to an exact weight freeze because projection and low-precision writeback remain active.
Our single-update intervention experiments explicitly distinguish these operations.

\section{Experimental Protocol}\label{sec:protocol}\subsection{Model, data, and training setup}

We train a custom 4B-parameter DeepSeekMoE-style language model from scratch, with 24 layers, hidden width 1,536, 128 routed experts of width 256 and top-2 routing, and one shared expert per layer. Its input and output embeddings are tied. The routed/shared design follows DeepSeekMoE~\cite{dai2024deepseekmoe}, with a separate traffic-driven router-bias controller~\cite{wang2024lossfree,deepseek2024v3}. Neither mechanism enters EnergyCal's coefficient.

Each run uses eight H100 80GB GPUs, 2,048-token sequences, and 131,072 tokens per optimizer update. A complete run has 10,000 updates (1.31B tokens), with 500 warmup updates followed by cosine decay to 10\% of peak learning rate. All methods use gradient clipping at norm 1, five NS iterations, and the same logical matrix partitions. These experiments do not claim compute-optimal or fully converged pretraining. Appendix~\ref{app:implementation} gives the relevant implementation details.

Training uses FineWeb \texttt{sample-10BT}~\cite{penedo2024fineweb}, with a deterministic train/validation partition. All final runs share the same data order; the independently varied initialization seeds do not constitute independently sampled corpora. Validation uses a fixed 524,288-token bank evaluated every 500 updates. Banks are fixed within each comparison but differ between development, Cohort H, and Cohort M. They are not certified previously unused external test sets across the complete exploratory history.

\subsection{Cohort H: development and confirmation}

Development screening evaluates the MuonH and plain-Muon learning-rate candidates over two initialization runs, using 3,000 updates of the 10,000-update schedule. Each baseline is selected by mean screening-endpoint validation cross-entropy (CE), not spike frequency. Both selected learning rates are 0.005, at the lower boundaries of their respective bounded search grids. The EnergyCal--MuonH recipe remains fixed at LR 0.01 and $p=1$; its settings are not reselected using confirmation outcomes.

\begin{table}[t]
\centering\small
\caption{Frozen Cohort H recipes. These are development-selected, best-tested settings rather than a coefficient-only factorial. Fallback peak LR is 0.001; weight decay applies only to eligible parameters.}
\label{tab:recipes}
\begin{tabular}{lrrr}
\toprule
Recipe & Matrix LR & Matrix WD & Fallback WD\\
\midrule
MuonH & 0.005 & 0 & 0\\
EnergyCal ($p=1$) & 0.010 & 0 & 0\\
Muon + WD & 0.005 & 0.1 & 0.1\\
\bottomrule
\end{tabular}
\end{table}

Confirmation compares these recipes over three new initialization runs for the complete schedule. The primary quality endpoint is final validation CE at fixed token budget. Conditions are paired by initialization and use matched recorded hardware and environments, although host and initialization effects are not fully crossed. Because learning rate differs between MuonH recipes, and the plain-Muon comparator also differs in decay and optimizer formulation, Cohort H estimates recipe-level outcomes rather than the isolated effect of the coefficient.

\subsection{Cohort M: plain-Muon extension}
\label{sec:muon_protocol}

A separate study compares plain Muon with EnergyCal exponents $p\in\{0.5,1\}$ over three new initialization runs. All nine runs complete 10,000 updates. Within each initialization, the methods use the same host and evaluation bank. All conditions share matrix LR 0.005, eligible matrix/fallback weight decay 0.1, and fallback LR 0.001. The sole optimizer change is the routed coefficient in Eq.~\eqref{eq:plain_energy}; the plain-Muon baseline is rerun under the same implementation as the calibrated variants. No learning-rate retuning or retrospective exponent selection is performed.

Cohort M uses a different evaluation bank and implementation revision from Cohort H. Their absolute CE/PPL values cannot be pooled or used to rank calibrated Muon against MuonH. Appendix~\ref{app:muon_extension} gives additional results.

\subsection{Stability and statistical reporting}

We quantify excursions using a fixed adaptive training-loss detector. For each update $t>500$, let $m_t$ and $a_t$ denote the median and median absolute deviation of the preceding 101 logged losses, excluding the current observation. The detection threshold is
\begin{align}
\tau_t&=\max\{m_t+0.5,\ 1.2m_t,\ m_t+8\sigma_t\},\\
\sigma_t&=\max\{1.4826a_t,10^{-6}\}.
\end{align}
At least 20 preceding observations are required. We flag $L_t>\tau_t$ and merge flags separated by at most 16 steps into an episode. The thresholded loss-spike burden is
\begin{equation}
B_1=\sum_{t:\,L_t>\tau_t,\ t>500}(L_t-m_t).
\label{eq:burden}
\end{equation}
It sums loss above the rolling median at flagged updates, not excess above the threshold or an integral over entire episodes. Its reduction is not a reduction in divergence probability. We also report episodes, maximum post-warmup training loss, and final validation CE, since persistent high loss can escape the adaptive detector.

Repeated-run outcomes are reported as arithmetic mean $\pm$ sample standard deviation (SD). Final quality comparisons use $n=3$ initialization runs per condition; development screening uses $n=2$ and timing uses three execution repeats. For $x_1,\ldots,x_n$,
\begin{equation}
\overline{x}=\frac{1}{n}\sum_{j=1}^{n}x_j,\qquad
s_x^2=\frac{\sum_{j=1}^{n}(x_j-\overline{x})^2}{n-1}.
\label{eq:sample_stats}
\end{equation}
The reported SD is descriptive, not a confidence interval; tokens, experts, and correlated steps are not independent replicates. Paired differences are computed within initialization before aggregation. Tabulated PPL uses per-run exponentiation; percentage reductions in geometric-mean PPL use $1-\exp(\overline{\Delta L})$. Burden percentages use ratios of condition-level mean burdens. The 95.98\% Cohort M summary is the equally weighted average of the two exponent-specific reductions, not a pooled six-run estimate. With three initializations per condition, no general significance inference is claimed.

\subsection{Timing and integrity}

Throughput is measured separately for Cohort H using three execution repeats per recipe on one eight-GPU node in cyclic method order. Each repeat runs 1,000 updates with evaluation, checkpointing, and heavy diagnostics disabled. Tokens divided by elapsed time over updates 501--1000 gives the run-level throughput; its arithmetic mean and SD are then reported across repeats. Peak allocated memory is maximized across recorded ranks within each run before aggregation. These short benchmarks do not measure equal-wall-time training quality.

All full-schedule runs finish. Recorded configurations, fixed validation banks, and selected state/input checks support within-comparison consistency, but not bitwise-identical trajectories or exhaustive data identity. Infrastructure failures and retries are not counted as optimizer divergence.

\section{Full-Schedule Results}\label{sec:results}
\subsection{Cohort H: recipe-level quality and stability}

Cohort H evaluates the frozen EnergyCal--MuonH recipe against development-selected MuonH and plain-Muon baselines over three confirmation seeds. 
As shown in Table~\ref{tab:main}, EnergyCal improves both final validation quality and training stability relative to MuonH. 
Mean final validation cross-entropy (CE) decreases from $3.2760\pm0.0270$ to $3.2345\pm0.0011$.
The paired CE difference is $-0.0415\pm0.0260$ nats/token, corresponding to 4.06\% lower geometric-mean perplexity (PPL).
All three paired endpoints favor EnergyCal.

\begin{table*}[t]
\centering\small
\caption{Cohort H confirmation: complete 10k schedule,
1.3107B tokens per run.
Values are arithmetic mean $\pm$ sample SD over $n=3$ unused initialization seeds. 
PPL is computed per run before aggregation; episodes and burden $B_1$ are per-run quantities.
$L_{\max}$ is the maximum post-warmup training loss within each run before aggregation. 
LRs are selected within bounded development grids.}
\label{tab:main}
\begin{tabular}{lrrrrrr}
\toprule
Recipe & Peak LR & Final CE & PPL & Episodes & $B_1$ & $L_{\max}$ \\
\midrule
MuonH & 0.005 & $3.2760 \pm 0.0270$ & $26.476 \pm 0.721$ & $37.33 \pm 4.04$ & $2046.01 \pm 1289.75$ & $39.99 \pm 12.64$ \\
EnergyCal ($p=1$) & 0.010 & $3.2345 \pm 0.0011$ & $25.394 \pm 0.027$ & $3.33 \pm 2.89$ & $10.11 \pm 9.65$ & $5.92 \pm 1.07$ \\
Muon + WD & 0.005 & $3.2410 \pm 0.0060$ & $25.560 \pm 0.152$ & $8.33 \pm 3.51$ & $67.41 \pm 52.76$ & $7.66 \pm 1.83$ \\
\bottomrule
\end{tabular}

\end{table*}

The largest improvement is in loss-spike burden.
Mean thresholded burden $B_1$ decreases from $2046.01\pm1289.75$ for MuonH to $10.11\pm9.65$ for EnergyCal, a 99.51\% reduction in the ratio of condition means. 
Mean spike episodes decrease from $37.33\pm4.04$ to $3.33\pm2.89$.
The maximum post-warmup training CE, summarized across runs, also falls from $39.99\pm12.64$
to $5.92\pm1.07$, supporting smaller excursions independently of the burden metric.
Both endpoint CE and burden improve in all three paired runs. 
No EnergyCal run triggers the detector between steps 4001 and 10,000, although this does not establish spike-free training.

Plain Muon with weight decay (Muon + WD) provides a stronger comparator. 
EnergyCal has a paired endpoint CE difference of $-0.0065\pm0.0057$ nats/token relative to Muon + WD, equivalent to 0.65\% lower geometric-mean PPL.
Its mean burden is 85.01\% lower, but the burden comparison is not favorable in every seed: plain Muon has lower burden in one of the three runs. 
These percentages are reductions in condition-level mean burdens, not averages of per-run reductions.

All compared recipes complete their training schedules. 
The results support an improved stability--quality trade-off for the tested EnergyCal--MuonH recipe relative to the development-selected MuonH baseline. 
Because the recipes differ in learning rate and, for plain Muon, weight decay and optimizer formulation, these comparisons do not isolate the effect of energy calibration or establish
a reduction in optimizer failure probability.

\subsection{Intermediate quality and the late crossover}

\begin{figure*}[t]
\centering
\includegraphics[width=.97\textwidth]{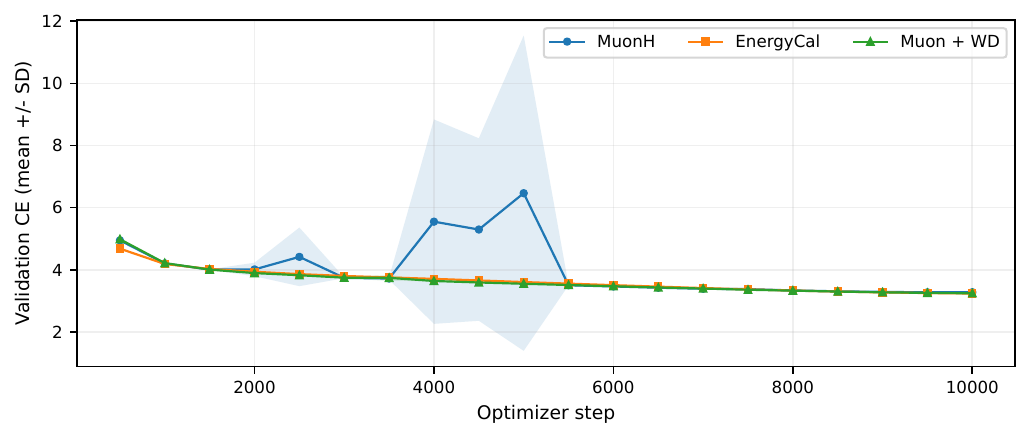}
\caption{Fixed-bank validation CE across Cohort H confirmation runs. 
Lines show means and shaded bands show $\pm1$ sample SD over three initialization seeds. 
All 20 evaluations are included, including loss excursions. 
Bands describe run-to-run variation, not confidence intervals. 
The predefined primary endpoint is step 10,000; development observations are not concatenated with these curves.}
\label{fig:curves}
\end{figure*}

EnergyCal's final quality advantage does not extend uniformly across training. 
Figure~\ref{fig:curves} shows all fixed-bank validation measurements, including evaluations affected by loss excursions. 
At step 3000, mean CE is 3.7995 for EnergyCal versus 3.7437 for Muon + WD; at step 5000, the corresponding values are 3.6038 and 3.5479.
The mean ordering reverses between the observed 8000- and 8500-step evaluations (supplementary Fig.~\ref{fig:late}).
Thus, EnergyCal's endpoint advantage emerges late in the schedule rather than persisting throughout training. 
Appendix~\ref{app:confirmation} gives the run-level summaries and late-window comparison.

To summarize intermediate quality, we compute normalized trapezoidal CE over the observed evaluations from steps 500 to 10,000.
The per-run summaries yield $3.6270\pm0.0016$ for EnergyCal, $3.6111\pm0.0094$ for Muon + WD, and $3.9923\pm0.3394$ for MuonH (mean $\pm$ sample SD).
By this measure, plain Muon retains better intermediate quality than EnergyCal, despite EnergyCal's better final endpoint.

This summary depends on evaluation spacing and the timing of individual spikes; it is not an integral of continuously measured validation loss. 
As a post hoc sensitivity check, averaging the evaluations at steps 9000, 9500, and 10,000 favors EnergyCal over both comparators in every paired run.
The predefined final endpoint remains the primary quality measure.

\subsection{Throughput, memory, and equal-time interpretation}

EnergyCal's stability improvements come with a modest throughput cost. 
Table~\ref{tab:benchmark} reports three execution repeats of each frozen Cohort H recipe.
EnergyCal achieves $65{,}976.8\pm353.3$ tokens/s, compared with $67{,}061.4\pm570.4$ for MuonH and $68{,}458.9\pm487.7$ for Muon + WD.
The ratios of mean throughput correspond to reductions of 1.62\% and 3.63\%, respectively, or increases of 1.64\% and 3.76\% in estimated fixed-token training time. 
Peak allocated memory is effectively unchanged relative to MuonH, consistent with the absence of additional persistent optimizer state.

\begin{table*}[t]
\centering\small
\caption{Cohort H production-style benchmark:
arithmetic mean $\pm$ sample SD over $n=3$ execution repeats on one eight-H100 node, not three quality seeds. 
Throughput is aggregated within each run over steps 501--1000. 
Optimizer time is summarized first by its within-run median and then across repeats. 
Peak allocated memory is first maximized over recorded ranks, including initialization, and then summarized across repeats. 
SD 0.0000 indicates identical recorded peaks, not an absence of hardware variability.}
\label{tab:benchmark}
\begin{tabular}{lrrr}
\toprule
Recipe & Tokens/s & Optimizer (s) & $M_{\rm alloc}$ (GiB) \\
\midrule
MuonH & $67061.4 \pm 570.4$ & $0.5460 \pm 0.0017$ & $39.7224 \pm 0.0000$ \\
EnergyCal ($p=1$) & $65976.8 \pm 353.3$ & $0.5487 \pm 0.0015$ & $39.7224 \pm 0.0000$ \\
Muon + WD & $68458.9 \pm 487.7$ & $0.4824 \pm 0.0014$ & $39.7245 \pm 0.0000$ \\
\bottomrule
\end{tabular}

\end{table*}

Measured throughput reflects the full training execution, including routing and optimizer-state differences, rather than only the computational cost of the calibration coefficient.
Moreover, these short benchmarks do not establish equal-wall-time training quality.

We did not conduct a full wall-clock-matched accuracy experiment. 
As an illustrative sensitivity check, we assume constant mean throughput and linearly interpolate the measured validation curves.
At the estimated time when plain Muon completes 10,000 steps, EnergyCal would reach approximately
9637 steps, with interpolated mean CE 3.2453, compared with plain Muon's observed 3.2410.
Under the same approximation, EnergyCal remains ahead of MuonH. 
These calculations are not measured time-to-quality results, and the small fixed-token endpoint advantage over plain Muon should not be interpreted as equal-time superiority.

\subsection{Cohort M: stabilization without Hyperball}
\label{sec:muon_transfer}

Cohort M tests whether EnergyCal also reduces loss excursions when applied to plain Muon, without Hyperball normalization or projection.
Unlike Cohort H, this study holds nominal learning rate and weight decay fixed. 
Table~\ref{tab:muon_extension} reports the results over three new initialization seeds.

Relative to its plain-Muon baseline, EnergyCal reduces mean thresholded loss-spike burden by 95.98\% on average across the two tested exponents. 
This is the equally weighted average of the two exponent-specific reductions in condition-level mean burden, not an average of per-seed percentage reductions (Appendix~\ref{app:muon_mean}).
Both exponents reduce burden in all three paired runs.

Endpoint improvements are smaller.
For $p=0.5$ and $p=1$, the paired CE differences relative to plain Muon are
$-0.0042\pm0.0042$ and $-0.0038\pm0.0028$ nats/token, respectively.
These correspond to 0.423\% and 0.380\% lower geometric-mean PPL.
The $p=0.5$ variant improves two of three paired endpoints, while $p=1$ improves all three. 
With only three seeds, these small quality differences remain descriptive.
The main finding is substantially reduced loss-spike burden without a large penalty in final validation quality.

\begin{table*}[t]
\centering\small
\caption{Cohort M: EnergyCal added to plain Muon without Hyperball. 
All conditions share matrix LR 0.005, eligible matrix and fallback weight decay 0.1, and the same 10k-update token budget. 
Values are mean $\pm$ sample SD over $n=3$ new initialization seeds; episodes and $B_1$ are per-run quantities. 
The validation bank differs from Cohort H, so absolute CE and PPL values must not be compared directly with Table~\ref{tab:main}.}
\label{tab:muon_extension}
\begin{tabular}{lrrrr}
\toprule
Recipe & Final CE & PPL & Episodes & $B_1$ \\
\midrule
Muon + WD & $3.189786 \pm 0.002381$ & $24.283 \pm 0.058$ & $11.67 \pm 4.04$ & $84.79 \pm 33.13$ \\
Muon + EnergyCal ($p=0.5$) & $3.185547 \pm 0.001877$ & $24.181 \pm 0.045$ & $3.00 \pm 2.00$ & $4.25 \pm 4.06$ \\
Muon + EnergyCal ($p=1$) & $3.185975 \pm 0.001173$ & $24.191 \pm 0.028$ & $1.67 \pm 1.15$ & $2.57 \pm 2.46$ \\
\bottomrule
\end{tabular}

\end{table*}

Neither tested exponent consistently dominates the other. 
The paired endpoint CE difference between $p=0.5$ and $p=1$ is only $-0.0004\pm0.0016$ nats/token.
We therefore report both settings without selecting an optimal exponent.
The observed-range mean validation CE is $3.5787\pm0.0428$ for plain Muon, $3.5490\pm0.0027$ for $p=0.5$, and $3.5508\pm0.0008$ for $p=1$.
A plain-Muon spike at the 4000-step evaluation contributes substantially to this difference.
Supplementary curves retain this observation and show that neither calibrated variant is best at every evaluation.

Coefficient telemetry provides additional context.
After averaging the recorded coefficients within each run, the mean coefficients across three initialization runs are
$0.9679\pm0.0021$ for $p=0.5$ and $0.9598\pm0.0017$ for $p=1$ (mean $\pm$ sample SD).
These summaries describe the applied attenuation, not its causal contribution to stabilization. Zero coefficients also occur, so not every routed direction is retained.
However, a routed-only uniform-scaling control has not been evaluated.
Consequently, matched nominal learning rate and weight decay do not establish whether the stability gains specifically require energy-dependent coefficient assignment or primarily reflect reduced routed-update magnitude.
Appendix~\ref{app:scalar_control} describes the missing control, and Section~\ref{sec:discussion} discusses the resulting limitations.

\section{Diagnostics and Local Interventions}\label{sec:diagnostics}\subsection{Energy, rank, and amplitude ablations}

A separate single-initialization audit compares six coefficient conditions at common matrix LR 0.01 over the first 3,000 updates (Table~\ref{tab:audit}). EnergyCal with $p=1$ has the lowest thresholded spike burden, whereas a power-iteration-based stable-rank coefficient has the lowest endpoint CE. The amplitude-only and near-zero-gate conditions do not match both outcomes. The amplitude-only run ends during a loss excursion; it is not labeled permanent divergence. These exploratory results are not extra confirmation seeds.

\begin{table}[t]
\centering\small
\caption{Single-initialization diagnostic audit at common matrix LR 0.01, over 3k updates. Alternative coefficients are not matched for effective update magnitude. An endpoint occurring during a spike is retained.}
\label{tab:audit}
\begin{tabular}{lrrr}
\toprule
Condition & 3k CE & Ep. & $B_1$ \\
\midrule
MuonH & 3.791280 & 4 & 430.51 \\
Energy ($p=0.5$) & 3.774685 & 4 & 89.03 \\
EnergyCal ($p=1$) & 3.769136 & 3 & 10.59 \\
Stable-rank & 3.761237 & 4 & 76.17 \\
Amplitude only & 9.318367 & 9 & 364.67 \\
Near-zero gate & 3.814775 & 6 & 531.36 \\
\bottomrule
\end{tabular}

\end{table}

The audit independently measures NS direction spectra and actual projected and BF16-stored parameter displacements using CPU float64 SVD. Sampled routed-down matrices include fixed experts and low-positive-energy selections; the sample is biased toward low-energy directions and is not an all-expert census. Crucially, a routed expert can receive zero current tokens yet retain momentum. One such NS direction has $\normF{D}=0.9921$, $\normop{D}=0.9724$, and $\sr(D)=1.0409$, illustrating that low stable rank does not imply small operator amplitude. Energy also depends on amplitude, as in Eq.~\eqref{eq:factorization}.

\begin{figure}[t]
\centering\includegraphics[width=\columnwidth]{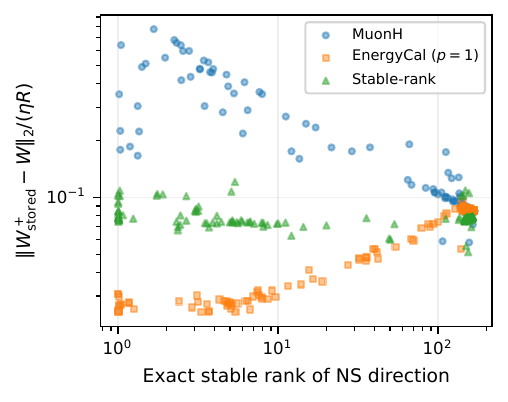}
\caption{Selected diagnostic measurements of direction stable rank and BF16-stored operator displacement. Stored movement is measured directly; the selected observations do not establish a general bound or a causal link to spikes.}
\label{fig:spectral}
\end{figure}

These measurements motivate studying post-NS energy but do not identify energy as a calibrated reliability estimate, select a universally optimal exponent, or show that Hyperball is the sole source of loss spikes. Appendix~\ref{app:interventions} specifies the ablations and interventions.

\subsection{Single-update interventions do not rescue two events}

We examine two baseline-scout loss excursions, events A and B. For each, candidate routed-down matrices were selected using low stable rank and traffic-weighted projected operator displacement, with same-bank controls. The heuristic was devised from exploratory captures but fixed before intervention outcomes. Seven branches per event modify one optimizer update and then return to the canonical training procedure (Appendix~\ref{app:interventions}). 
These events were selected for diagnostic analysis and should not be treated as representative samples of loss spikes.

All 14 branches complete. Initial state and recorded input checks agree within each event, although replays are not certified bitwise-identical. Exact weight freezing restores the original parameter bits while retaining momentum and ordinary writeback behavior; it does not remove an expert.

\begin{table}[t]
\centering\small
\caption{Next-step training CE after single-update interventions in two distinct saved-state events. Pre-intervention CE is 4.1928 (A) and 4.2143 (B). No tested branch eliminates the subsequent loss rise.}
\label{tab:local}
\begin{tabular}{lrr}
\toprule
Single-update intervention & Event A & Event B \\
\midrule
Canonical & 6.3165 & 6.9069 \\
Candidate: freeze & 6.3172 & 6.9110 \\
Control: freeze & 6.3160 & 6.9085 \\
All routed: EnergyCal & 6.3024 & 6.9873 \\
\bottomrule
\end{tabular}

\end{table}

Candidate-only EnergyCal reduces actual BF16-stored operator displacement by 86.3\% (A) and 76.3\% (B), while exact freezing eliminates that candidate displacement. Nevertheless, next-step CE remains near 6.3 and 6.9, respectively. In the common short windows, candidate freezing reduces burden by 3.09\% and 0.76\%, compared with 1.61\% and 1.40\% for control freezing. Applying EnergyCal to all routed matrices for a single update reduces short-window burden by 2.12\% in event A but increases it by 3.41\% in event B. This \emph{short-window burden} is distinct from the full-schedule $B_1$ defined in Eq.~\eqref{eq:burden}.

A separate fixed-input 512-token probe shows no large reduction in output KL; probe CE decreases slightly in every branch. 
The probe differs from the next high-loss training batch and omits the subsequent router-bias update.
These interventions therefore do not explain continuous calibration's observed full-schedule benefit through immediate removal of the two examined spikes, but they do not establish that smaller or history-dependent effects are absent.

\section{Discussion and Limitations}\label{sec:discussion}
\leadpara{Empirical contribution and practical trade-offs.}
The main finding is a substantial reduction in thresholded loss-spike burden with both MuonH and plain Muon, accompanied by small improvements in final validation quality. 
In Cohort H, the frozen EnergyCal--MuonH recipe reduces mean burden by 99.51\% relative to a development-selected, lower-rate MuonH recipe, with improved endpoint quality and 1.62\% lower mean
throughput. 
Because the learning rates differ, this comparison does not isolate the effect of calibration alone. 
The plain-Muon comparator is faster and achieves better intermediate quality; an illustrative equal-time estimate can also favor plain Muon.
In Cohort M, EnergyCal reduces mean burden by 95.98\% on average across two tested exponents at matched nominal learning rate and weight decay, showing that stabilization is also observed without Hyperball. 
However, different validation banks preclude a direct quality ranking of the calibrated optimizer families, and no production-style throughput benchmark is available for Cohort M.
These results support empirical stabilization in the evaluated setting, not general optimizer superiority.

\leadpara{The unresolved uniform-scaling alternative.}
EnergyCal attenuates routed-expert updates, but the observed improvements need not arise specifically from assigning different coefficients according to direction energy.
Cohort M lacks a routed-only uniform-scaling or update-budget-matched control, so it cannot distinguish energy-dependent attenuation from the benefit of reducing routed-update magnitudes more generally. 
A fixed coefficient near 0.96, leaving weight decay and non-routed updates unchanged, would provide an informative comparison, although it would not exactly match the update-norm budget.
Appendix~\ref{app:scalar_control} distinguishes these controls.
Furthermore, we do not establish relative superiority over other expert-wise scaling methods, including Compass and Excitation, because native-method head-to-head comparisons were not performed.

\leadpara{Geometry and causal interpretation.}
The geometric analysis provides motivation for attenuation, not a causal explanation of loss spikes. 
The inverse-stable-rank identity describes the operator norm of MuonH's trial update, not the final projected or stored parameter displacement, and does not directly apply to plain Muon.
Moreover, the energy statistic depends on both spectral concentration and direction amplitude, rather than stable rank alone.
In two examined events, single-update interventions substantially reduce the nominated parameter movements but fail to prevent the subsequent loss spikes.
Thus, the full-schedule benefits cannot be explained simply as immediate removal of these two events through one-step attenuation. 
Effects accumulated over training, including changes in expert parameters or routing dynamics, remain possible explanations but are not established here.
These negative interventions limit the mechanistic interpretation without negating the observed benefits of continuous calibration.

\leadpara{Scope, measurement, and generalizability.}
Both cohorts use one custom 4B-parameter, 128-expert, top-2 MoE model and 1.31B training tokens per run, with three initialization seeds and a shared data order.
The results therefore do not establish generalization across architectures, scales, data distributions, or training budgets. 
Cohort H selects baselines within bounded learning-rate grids whose best-tested values lie at the lower boundaries; Cohort M fixes the plain-Muon recipe rather than independently retuning every variant.
Reported standard deviations describe run-to-run variation, not confidence intervals. The adaptive spike detector can miss persistent loss elevation, while periodic validation may miss transient quality changes. 
Selected spectral measurements are not an all-layer census, and the local probes omit the subsequent router-bias-controller update.

\section{Conclusion}\label{sec:conclusion}EnergyCal applies a simple, one-sided attenuation to routed-expert updates using finite-step Newton–Schulz direction energy.
In a 4B-parameter sparse MoE model, it substantially reduces thresholded loss-spike burden with both MuonH and plain Muon, showing that stabilization is also observed without Hyperball in the evaluated setting.
The frozen EnergyCal–MuonH recipe reduces mean burden by 99.51\% and final geometric-mean perplexity by 4.06\% relative to a development-selected, lower-rate MuonH recipe, with 1.62\% lower mean throughput.
In a separate plain-Muon study at matched nominal learning rate and weight decay, reductions in mean burden average 95.98\% across the two tested exponents, with small endpoint gains. 
These results support empirical stabilization in the evaluated MoE setting rather than general optimizer superiority. 
However, single-update interventions did not rescue two examined spikes, and the absence of a routed-only uniform-scaling control leaves the specific contribution of energy-dependent attenuation unresolved. 
Further controlled comparisons are needed to distinguish this contribution from reduced update magnitude.

\FloatBarrier
\clearpage
{\small\sloppy\bibliographystyle{IEEEbib}\bibliography{energycal_references}}
\clearpage
\appendix
\section{Implementation and Evaluation Details}
\label{app:implementation}

This appendix records the numerical and parameter-partition details needed to interpret EnergyCal. The model, training budget, and comparison design are described in Section~\ref{sec:protocol}.

\subsection{Finite-step Newton--Schulz}
The matrix input to NS is cast to BF16 and, if needed, transposed so that the smaller dimension defines the Gram matrix. For the working matrix $X$, normalization and five quintic iterations are
\begin{equation}
X\leftarrow X/\max(\normF{X},10^{-7}),
\end{equation}
\begin{align}
Q&=XX^\top,\\
X&\leftarrow3.4445X+(-4.7750Q+2.0315Q^2)X.
\end{align}
The norm is accumulated in FP32, and the output is returned in FP32 with its original orientation. Finite-step BF16 NS does not guarantee exact unit singular values, so post-NS direction energy remains informative. MuonH separately uses a denominator guard $\epsilon_H=10^{-12}$ for normalization and projection.

\subsection{Logical matrices and writeback}
Fused gate/up storage is split into separate logical matrices for optimization; each routed gate, up, or down matrix has $d=\min(m,n)=256$. EnergyCal acts only on routed matrices. Attention and shared-expert matrices retain their matrix-optimizer updates, and embeddings, router weights, and vectors use an Adam-style fallback.

MuonH uses a fixed initial Frobenius radius and the trial update and projection of Eqs.~\eqref{eq:trial}--\eqref{eq:project}. Plain Muon has no sphere projection and uses Eq.~\eqref{eq:plain_energy}. Both use a common dithered BF16 writeback policy. Thus $q=0$ suppresses the NS trial but need not produce an exact bitwise freeze: projection and writeback may change MuonH weights, while plain Muon can still apply weight decay.

\subsection{Geometry and coefficient interpretation}
For nonzero $D$, $\sr(D)=\normF{D}^2/\normop{D}^2$. With the MuonH denominator guard, the trial operator norm is
\begin{equation}
\normop{\Delta^{\rm trial}}=\frac{\eta Rq\,\normop{D}}{\max(\normF{D},\epsilon_H)}.
\end{equation}
Equation~\eqref{eq:identity} follows when the guard is inactive. Stable rank is undefined at the zero matrix.

For comparison, the unclipped, unguarded exact-stable-rank coefficient $q_r=\sqrt{\sr(D)/(d\rho_{\rm ref})}$ gives
\begin{equation}
\normop{\Delta^{\rm trial}}=\frac{\eta R}{\sqrt{d\rho_{\rm ref}}}.
\end{equation}
The tested stable-rank alternative estimates operator norm rather than computing an exact SVD during training. In contrast, EnergyCal with $p=1/2$ yields $\normop{\Delta^{\rm trial}}=\eta R\normop{D}/\sqrt{d\rho_{\rm ref}}$ in the same regime. These identities characterize the trial, not the projected/stored displacement or a bound on the network's loss change.

\subsection{Validation boundaries}
A deterministic partition of FineWeb \texttt{sample-10BT} reserves 0.5\% of rows for validation. Evaluation uses fixed banks within each comparison, but development, Cohort H, and Cohort M use different banks. Different banks and implementation revisions prevent direct cross-cohort ranking of absolute CE/PPL. The validation observations are not certified as previously unused external test examples, and content-level near-duplicate contamination was not independently audited.

\section{Development Selection and Confirmation Detail}
\label{app:confirmation}

\subsection{Development screening}
The baseline learning-rate candidates are screened over two initialization runs, each observed after 3,000 updates. The selected baseline rate minimizes mean screening-endpoint validation CE. Both baseline choices are at the lower boundaries of their tested grids, so the comparison is best-tested within a bounded search. EnergyCal--MuonH is fixed at matrix LR 0.01, $p=1$, $\rho_{\rm ref}=0.8$, and $q_{\min}=0$ before confirmation.

\begin{table*}[t]
\centering\small
\caption{Completed development screen: mean $\pm$ sample SD over $n=2$ initialization runs after 3,000 updates. All endpoints, including unfavorable ones, are retained. Episodes and burden are per-run quantities.}
\label{tab:screen}
\begin{tabular}{lrrrr}
\toprule
Recipe & LR & Final CE & Episodes & $B_1$ \\
\midrule
MuonH & 0.005000 & $3.703740 \pm 0.013658$ & $8.00 \pm 0.00$ & $339.40 \pm 379.07$ \\
MuonH & 0.007071 & $3.733119 \pm 0.013795$ & $7.00 \pm 1.41$ & $562.76 \pm 0.88$ \\
MuonH & 0.010000 & $3.949324 \pm 0.226689$ & $11.50 \pm 4.95$ & $592.87 \pm 99.21$ \\
EnergyCal ($p=1$) & 0.010000 & $3.769425 \pm 0.000647$ & $2.50 \pm 0.71$ & $8.76 \pm 0.24$ \\
Muon + WD & 0.005000 & $3.703061 \pm 0.004366$ & $1.50 \pm 0.71$ & $1.84 \pm 1.53$ \\
Muon + WD & 0.010000 & $3.772394 \pm 0.011707$ & $5.50 \pm 0.71$ & $75.87 \pm 0.14$ \\
Muon + WD & 0.020000 & $3.909822 \pm 0.017177$ & $5.00 \pm 1.41$ & $69.51 \pm 12.91$ \\
\bottomrule
\end{tabular}

\end{table*}

The selected lower-rate baselines have better development endpoints than EnergyCal. The independent full-schedule confirmation therefore tests a previously fixed recipe, not an extrapolation of the shorter screen. Development and confirmation validation banks are not combined.

\subsection{Confirmation effects}
\begin{table*}[t]
\centering\small
\caption{Paired EnergyCal-minus-baseline differences over $n=3$ confirmation initializations (mean $\pm$ sample SD). Negative values favor EnergyCal. Pairing precedes aggregation.}
\label{tab:seed_detail}
\begin{tabular}{lrrr}
\toprule
EnergyCal minus & CE difference & $B_1$ difference & Episode difference \\
\midrule
MuonH & $-0.041491 \pm 0.026043$ & $-2035.91 \pm 1297.37$ & $-34.00 \pm 1.73$ \\
Muon + WD & $-0.006510 \pm 0.005709$ & $-57.31 \pm 55.57$ & $-5.00 \pm 6.24$ \\
\bottomrule
\end{tabular}

\end{table*}

Percentage burden reductions use $1-\overline B_E/\overline B_C$, with $\overline B$ denoting the condition-level mean burden. This gives 99.51\% versus MuonH and 85.01\% versus Muon + WD; neither is the average of run-wise percentage reductions. All confirmation runs contribute, including the zero-burden EnergyCal run and the unfavorable burden comparison against Muon + WD.

\subsection{Intermediate-quality summaries}
For evaluation steps $s_j\in\{500,1000,\ldots,10000\}$, normalized trapezoidal CE is computed per run:
\begin{equation}
\overline L_{\rm obs}=\frac{1}{s_K-s_1}\sum_{j=1}^{K-1}(s_{j+1}-s_j)\frac{L_{s_j}+L_{s_{j+1}}}{2},\qquad K=20.
\end{equation}
Mean and sample SD are then computed across runs. A separate post hoc late-window check averages the last three evaluations within each run. Neither summary replaces the predefined final endpoint. The observed-range summary favors plain Muon over EnergyCal, whereas the late-window summary favors EnergyCal.

\begin{table*}[t]
\centering\small
\caption{Cohort H curve summaries: mean $\pm$ sample SD of run-level results. The late-window comparison is post hoc.}
\label{tab:curve_summaries}
\begin{tabular}{lrr}
\toprule
Recipe & Observed-range CE & Late-window CE \\
\midrule
MuonH & $3.992289 \pm 0.339400$ & $3.278503 \pm 0.007961$ \\
EnergyCal ($p=1$) & $3.627002 \pm 0.001572$ & $3.251435 \pm 0.000639$ \\
Muon + WD & $3.611137 \pm 0.009433$ & $3.256059 \pm 0.005045$ \\
\bottomrule
\end{tabular}

\end{table*}

\begin{figure}[t]
\centering\includegraphics[width=\columnwidth]{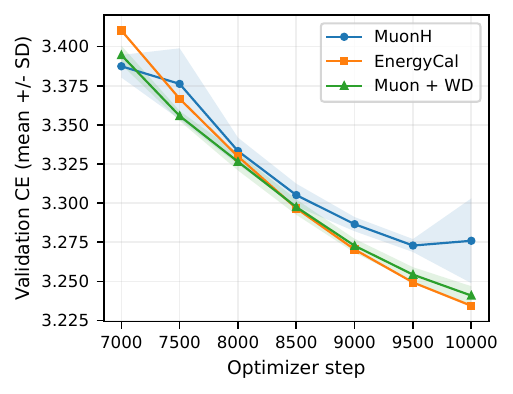}
\caption{Late-schedule validation CE (mean $\pm$ sample SD, $n=3$). EnergyCal's small endpoint advantage over plain Muon appears late; the full trajectory is shown in Fig.~\ref{fig:curves}.}
\label{fig:late}
\end{figure}

The adaptive spike detector may miss persistent high-loss periods, while narrow excursions can produce large maxima but modest burden. Final validation CE, maximum training loss, episodes, and thresholded burden therefore remain complementary outcomes. All final runs complete.

\section{Plain-Muon Transfer}
\label{app:muon_extension}

Cohort M tests EnergyCal without Hyperball. Plain Muon and its $p\in\{0.5,1\}$ EnergyCal variants are rerun over three new initializations for the complete 10,000-update schedule. The same architecture, data order, and token budget are used as in Cohort H, but a separate implementation revision and validation bank preclude direct cross-cohort ranking.

All Cohort M conditions share matrix LR 0.005, eligible matrix/fallback weight decay 0.1, and fallback peak LR 0.001. Only the routed coefficient in Eq.~\eqref{eq:plain_energy} changes; momentum, decay, non-routed updates, and the Adam fallback do not. EnergyCal uses $\rho_{\rm ref}=0.8$, $q_{\min}=0$, and the two stated exponents. Every run completes.

\begin{figure*}[t]
\centering\includegraphics[width=.97\textwidth]{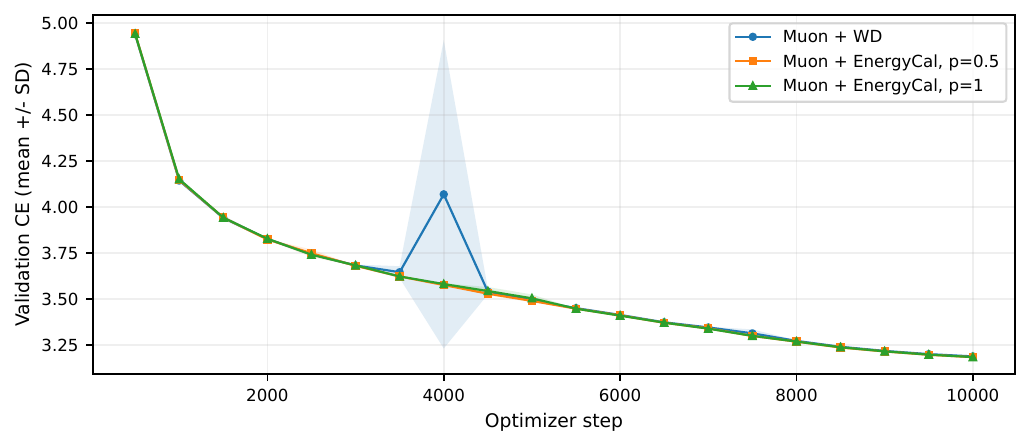}
\caption{Cohort M validation CE: mean $\pm$ sample SD over three runs on its own fixed evaluation bank. All observations, including the plain-Muon loss excursion at step 4000, are retained. Bands are not confidence intervals.}
\label{fig:muon_extension_curve}
\end{figure*}

\subsection{Quality and stability}
\label{app:muon_mean}
\begin{table*}[t]
\centering\small
\caption{Additional Cohort M results: mean $\pm$ sample SD over $n=3$ initialization runs. Observed-range CE integrates the 500--10,000 evaluations; late-window CE is a post hoc check over the final three evaluations.}
\label{tab:muon_extra}
\begin{tabular}{lrrrr}
\toprule
Recipe & PPL & $L_{\max}$ & Observed-range CE & Late-window CE \\
\midrule
Muon + WD & $24.2833 \pm 0.0578$ & $8.54 \pm 0.82$ & $3.578701 \pm 0.042809$ & $3.203759 \pm 0.002122$ \\
Muon + EnergyCal ($p=0.5$) & $24.1805 \pm 0.0454$ & $5.61 \pm 0.87$ & $3.548957 \pm 0.002669$ & $3.200299 \pm 0.001403$ \\
Muon + EnergyCal ($p=1$) & $24.1909 \pm 0.0284$ & $5.49 \pm 0.75$ & $3.550821 \pm 0.000788$ & $3.200778 \pm 0.001217$ \\
\bottomrule
\end{tabular}

\end{table*}

With $\overline B_p$ denoting the condition-level three-run mean burden for exponent $p$ and $\overline B_0$ the plain-Muon mean, the reduction is $r_p=100(1-\overline B_p/\overline B_0)$. The two reductions are 94.99\% and 96.97\%; their equal-weight average is 95.98\%. This is not a mean of per-initialization percentages or an estimate using six independent replicates.

\begin{table}[t]
\centering\small
\caption{Cohort M paired effects: mean $\pm$ sample SD over three within-initialization differences (EnergyCal minus plain Muon). Negative values favor calibration.}
\label{tab:muon_paired}
\begin{tabular}{lrr}
\toprule
Coefficient & $\Delta$ CE & $\Delta B_1$ \\
\midrule
$p=0.5$ & $-0.004239 \pm 0.004194$ & $-80.54 \pm 36.44$ \\
$p=1$ & $-0.003811 \pm 0.002849$ & $-82.23 \pm 30.67$ \\
\bottomrule
\end{tabular}

\end{table}

Both exponents reduce thresholded burden in all three paired runs. Endpoint CE improves in two of three runs for $p=0.5$ and all three for $p=1$; neither exponent is uniformly optimal. The paired endpoint CE difference between the calibrated variants is only $-0.0004\pm0.0016$ nats/token.

The mean observed-range advantage is partly influenced by a plain-Muon spike at one evaluation. Calibrated variants do not dominate at every point; the late-window check also remains post hoc. Final CE at 10,000 updates is the primary endpoint.

\subsection{Coefficient measurements}
Recorded coefficients are summarized within each run before aggregation across three initializations. Their means are $0.9679\pm0.0021$ for $p=0.5$ and $0.9598\pm0.0017$ for $p=1$ (mean $\pm$ sample SD). Some coefficients equal zero. Thus positive scaling preserves a nonzero NS direction, whereas $q=0$ suppresses that direction without necessarily freezing the stored parameter. These summaries are not activation-weighted effective learning rates, and no diagnostic-off production-style throughput benchmark was performed for this cohort.

\subsection{Unperformed routed-only scalar controls}
\label{app:scalar_control}
The matched nominal LR/decay comparison measures the effect of adding the evaluated attenuation rule but does not identify whether expert-wise energy-dependent assignment is better than uniform routed-update reduction. A simple untested control would fix $q_e=\alpha$ for all routed matrices (e.g., $\alpha=0.96$), leaving other updates and decay unchanged.

Equal average coefficients do not imply equal update-norm budgets. At a common state, let $U_{e,t}=-\eta_t s_eD_{e,t}$ denote the pre-decay, pre-writeback plain-Muon direction step for routed matrix $e$. When the denominator is positive, a norm-budget-matched uniform coefficient would be
\begin{equation}
\alpha_t=\left(\frac{\sum_{e\in\mathcal R}q_{e,t}^2\normF{U_{e,t}}^2}{\sum_{e\in\mathcal R}\normF{U_{e,t}}^2}\right)^{1/2}.
\label{eq:scalar_budget_control}
\end{equation}
This matches only the summed squared Frobenius norms of the direction steps at that state. It does not match stored parameter changes, decay, or future trajectories. Neither fixed-scalar nor norm-budget-matched controls were evaluated, so a specific causal advantage of assigning coefficients by energy remains unresolved.

\section{Spectral Audit and Event Protocol}
\label{app:interventions}

\subsection{Alternative coefficients and spectral measurements}
A separate single-initialization audit compares six update rules at common matrix LR 0.01 over 3,000 updates, with additional scouts used only for diagnostic observations. The stable-rank, amplitude-only, and near-zero-gate controls use
\begin{align}
q_{\rm rank}&=\min\left\{1,\sqrt{\frac{\widehat{\sr}(D)}{0.8d}}\right\},\\
q_{\rm amp}&=\min\{1,\widehat{\normop{D}}\},\\
q_{\rm zero}&=\mathbf1\{\normF{D}>10^{-7}\}.
\end{align}
The spectral quantities used during optimization are estimated by power iteration rather than exact SVD. These alternatives were not independently tuned to equal update budgets.

For diagnostics, NS spectra and projected/BF16-stored displacements are measured independently using CPU float64 SVD on selected routed-down matrices. Selection deliberately includes low-energy directions, so the sampled distribution is not an unbiased census. Norms reconstructed from the trial formula do not independently verify Eq.~\eqref{eq:identity}. A zero directional coefficient can still yield stored changes after projection and writeback, motivating exact-weight-freeze interventions.

\subsection{Single-update interventions}
Two saved-state spike events (A and B) are examined separately. Candidates are selected from routed-down matrices using low stable rank and traffic-weighted projected displacement, with comparable same-bank control matrices. This heuristic was informed by exploratory captures but fixed before treatment outcomes. Each event uses seven single-update branches: canonical update, candidate EnergyCal, candidate zero trial, candidate exact weight freeze, control freeze, all-routed EnergyCal, and all-routed EnergyCal with the candidate restored to canonical. Training then resumes normally.

Exact freezing restores the selected weight bits after the ordinary update and writeback while retaining momentum evolution and RNG consumption. It differs from zero trial, which suppresses only the direction proposal. Branches start from matched saved states and input sequences, but independent recomputation does not guarantee identical numerical trajectories.

\begin{table}[!t]
\centering\scriptsize
\setlength{\tabcolsep}{2.2pt}
\caption{All intervention branches in events A and B ($n=1$ per branch and event). $r_A,r_B$ are short-window burden improvements, not reductions in the full-schedule $B_1$.}
\label{tab:intervention_all}
\begin{tabular}{lrrrr}
\toprule
Intervention & CE A & CE B & $r_A$ (\%) & $r_B$ (\%) \\
\midrule
Canonical & 6.3165 & 6.9069 & 0.00 & 0.00 \\
Candidate ECal & 6.3163 & 6.9068 & +0.93 & +0.22 \\
Candidate zero & 6.3160 & 6.9109 & -1.25 & +0.32 \\
Candidate freeze & 6.3172 & 6.9110 & +3.09 & +0.76 \\
Control freeze & 6.3160 & 6.9085 & +1.61 & +1.40 \\
All routed ECal & 6.3024 & 6.9873 & +2.12 & -3.41 \\
All except cand. & 6.3020 & 6.9840 & +1.48 & -3.15 \\
\bottomrule
\end{tabular}

\end{table}

For an event's common short window $\mathcal W$ and fixed pre-intervention reference loss $b$, the local burden and percent improvement relative to its canonical continuation are
\begin{align}
A_{\rm win}&=\sum_{s\in\mathcal W}\max(0,L_s-b),\\
r_{\rm win}&=100\left(1-\frac{A_{\rm arm}}{A_{\rm canonical}}\right).
\end{align}
Unlike $B_1$, $A_{\rm win}$ includes all positive deviations above the fixed reference in the short window, without threshold-flag filtering. It is used only for intervention comparisons.
For events A and B, the evaluation windows are steps 1068--1100 and 926--957, with fixed reference losses $b=4.133466$ and $b=4.240423$, respectively.

Candidate-only EnergyCal reduces stored operator displacement by 86.3\% in event A and 76.3\% in event B, while exact freezing eliminates that displacement. The next-step loss rise persists. Candidate freeze reduces short-window burden by 3.09\% and 0.76\%, versus 1.61\% and 1.40\% for control freeze; all-routed EnergyCal reduces short-window burden by 2.12\% in event A but increases it by 3.41\% in event B. These small, inconsistent effects do not establish substantial immediate rescue or a zero causal effect.

\subsection{Fixed-input probes and causal limitations}
A fixed 512-token probe compares CE and output KL immediately before and after each treated update. Mean probe CE decreases slightly in every branch, with no large KL reduction from the tested corrections. The probe is different from the next high-loss training batch and omits the subsequent router-bias-controller update. Consequently, the local observations do not identify the combination of history, routing, and inputs responsible for the two events. They also do not negate the full-schedule improvement under continuous EnergyCal.

\end{document}